\documentclass{article}
\usepackage[nonatbib, preprint]{neurips_2024}

\usepackage[utf8]{inputenc} 
\usepackage[T1]{fontenc}    
\usepackage{hyperref}       
\usepackage{url}            
\usepackage{booktabs}       
\usepackage{amsfonts}       
\usepackage{microtype}      
\usepackage{xcolor}         
\usepackage{svg}

\usepackage{tikz}
\usepackage{pgfplots}
\pgfplotsset{compat=1.18}

\usepackage{amsmath}
\usepackage{amsfonts}
\usepackage{amssymb}
\usepackage{ mathrsfs }
\usepackage{graphicx}
\usepackage{amsthm}
\usepackage{booktabs}

\newenvironment{thisnote}{\par\color{blue}}{\par}

\usepackage[T1]{fontenc}
\usepackage{lmodern} 

\title{Geometric sparsification in recurrent neural networks}
\author{%
  Wyatt Mackey
    \\
  Department of Mathematics / NIMBioS \\ 
  University of Tennessee\\
  Knoxville, TN 37996 \\
  \texttt{wmackey2@utk.edu} \\
  \And 
  Ioannis Schizas \\ 
  DEVCOM ARL \\
  Army Research Lab \\
  Aberdeen, MD 21001 \\
  \texttt{ioannis.d.schizas.civ@army.mil} \\
  \And 
  Jared Deighton \\ 
  Department of Mathematics \\
  University of Tennessee \\ 
  Knoxville, TN 37996 \\ 
  \texttt{jdeighto@vols.utk.edu}
  \And
  David L. Boothe, Jr. \\
  DEVCOM ARL \\
  Army Research Lab \\
  Aberdeen, MD 21001 \\
  \texttt{david.l.boothe7.civ@army.mil} \\
  \And 
  Vasileios Maroulas \\ 
  Department of Mathematics \\
  University of Tennessee \\
  Knoxville, TN 37996 \\ 
  \texttt{vmaroula@utk.edu}
}
\date{}
\newtheorem{lem}{Lemma}[section]

\theoremstyle{definition}
\newtheorem{defn}[lem]{Definition}

\newcommand\nc{\newcommand}
\nc{\on}{\operatorname}
\nc\renc{\renewcommand}
\nc{\BR}{\mathbb R}
\nc{\BC}{\mathbb C}
\nc{\BQ}{\mathbb Q}
\nc{\BZ}{\mathbb Z}
\nc{\BN}{\mathbb N}
\nc{\BS}{\mathbb S}
\nc{\BA}{\mathbb A}
\nc{\holim}{\on{holim}}
\nc{\cS}{\mathcal S}
\nc{\cD}{\mathcal D}
\nc{\cE}{\mathcal E}
\nc{\BP}{\mathbb P}
\nc{\Hom}{\on{Hom}}
\nc{\wt}{\widetilde}
\nc{\vspan}{\on{span}}
\nc{\ord}{\on{ord}}
\nc{\im}{\on{im}}
\nc{\Mat}{\on{Mat}}
\nc{\can}{\on{can}}
\nc{\coker}{\on{coker}}
\nc{\ev}{\on{ev}}
\nc{\Tr}{\on{Tr}}
\nc{\End}{\on{End}}
\nc{\swap}{\on{swap}}
\nc{\Set}{\on{Set}}
\nc{\bC}{{\mathbf C}}
\nc{\bc}{{\mathbf c}}
\nc{\bD}{{\mathbf D}}
\nc{\bd}{{\mathbf d}}
\nc{\bE}{{\mathbf E}}
\nc{\be}{{\mathbf e}}
\nc{\bF}{{\mathbf F}}
\nc{\bff}{{\mathbf f}}
\nc{\CE}{\mathcal E}
\nc{\CO}{\mathcal O}
\nc{\CC}{\mathcal C}
\nc{\CM}{\mathcal M}
\nc{\CA}{\mathcal A}
\nc{\SC}{\mathscr C}
\nc{\SA}{\mathscr A}
\nc{\SB}{\mathscr B}

\nc{\adj}{\on{adj}}
\nc{\tensor}[3]{#1 \underset{#2}\otimes #3}
\nc{\Nat}{\on{Nat}}
\nc{\op}{\on{op}}
\nc{\Funct}{\on{Funct}}
\nc{\Ob}{\on{Ob}}
\nc{\fR}{\mathfrak{R}}
\nc{\Vect}{\on{Vect}}
\nc{\ns}{\on{non-spec}}
\nc{\ol}{\overline}
\nc{\ul}{\underline}
\nc{\univ}{\on{univ}}
\nc{\Maps}{\on{Maps}}
\nc{\bdd}{\on{bdd}}
\nc{\cont}{\on{cont}}
\nc{\Sym}{\on{Sym}}
\nc{\vol}{\on{vol}}
\nc{\supp}{\on{supp}}
\nc{\Lie}{\on{Lie}}
\nc{\master}{\on{master}}
\nc{\pt}{\on{pt}}
\nc{\bcd}{\[ \begin{tikzcd}}
\nc{\ecd}{\end{tikzcd} \]}
\nc{\funcon}{\on{Funct}_{\on{cont}}}
\nc{\funconpw}{\on{Funct}_{\on{cont},\on{pw-lin}}}
\nc{\ts}{\textsc}
\nc{\codim}{\on{codim}}
\nc{\fm}{\mathfrak m}
\nc{\SR}{\mathscr R}
\nc{\Tor}{\on{Tor}}
\nc{\Ext}{\on{Ext}}
\renc{\sc}[1]{\textsc{#1}}

\usepackage{caption}
\usepackage{subcaption}

\begin{document}

\maketitle

\begin{abstract}
A common technique for ameliorating the computational costs of running large neural models is \textit{sparsification}, or the pruning of neural connections during training.
Sparse models are capable of maintaining the high accuracy of state of the art models, while functioning at the cost of more parsimonious models. The structures which underlie sparse architectures are, however, poorly understood and not consistent between differently trained models and sparsification schemes. In this paper, we propose a new technique for sparsification of recurrent neural nets (RNNs), called \textit{moduli regularization}, in combination with magnitude pruning. Moduli regularization leverages the dynamical system induced by the recurrent structure to induce a geometric relationship between neurons in the hidden state of the RNN. By making our regularizing term explicitly geometric, we provide the first, to our knowledge, \textit{a priori} description of the desired sparse architecture of our neural net, as well as explicit end-to-end learning of RNN geometry. We verify the effectiveness of our scheme under diverse conditions, testing in navigation, natural language processing, and addition RNNs. Navigation is a structurally geometric task, for which there are known moduli spaces, and we show that regularization can be used to reach 90\% sparsity while maintaining model performance only when coefficients are chosen in accordance with a suitable moduli space. Natural language processing and addition, however, have no known moduli space in which computations are performed. Nevertheless, we show that moduli regularization induces more stable recurrent neural nets, and achieves high fidelity models above 90\% sparsity.
    \end{abstract}

\section{Introduction}
Sparsification, or the excision of neural connections during training, is an important technique in the manufacture of computationally efficient deep learning models. Neural nets used across applications are heavily over-parameterized, and therefore both resilient to substantial sparsification and prone to over-fitting \cite{frankle2018lottery, gale2019state}. Regularization and pruning techniques have therefore found great success in improving the quality of models trained, while simultaneously decreasing the computational costs of training and running large neural architectures. 

While model sparsification has been attempted using many different heuristics \cite{NEURIPS2022_sparsity, NEURIPS2022_089b592c_controlledsparsity, louizos2017learning, narang2017exploring, singh2020woodfisher}, the underlying sparse architectures are poorly understood, and vary greatly subject to small changes in both the sequence that training data is presented \cite{frankle2020linear} and the initial randomization of weights \cite{frankle2018lottery, gale2019state}. Frankle, et al.\ (2018) proposed the Lottery Ticket Hypothesis as a potential explanation of the instability of sparse models \cite{frankle2018lottery}. The Lottery Ticket Hypothesis suggests that sparse models arise from fortunate weight initializations on small subnetworks. Neural nets are designed to be heavily over-parameterized because this increases the probability of fortunate weight initializations on subnetworks, i.e.\ that some subnetwork ``wins the lottery." 

This hypothesis is intuitive and appealing. However, it leaves open the problem of describing sparse neural architectures. Retraining neural nets on the so-called ``winning lottery tickets" with re-randomized weights fails to recover high-quality networks in both Resnets and transformers \cite{gale2019state}. This presents fundamental obstacles to any project of structured \emph{ab initio} sparsity, though it does not rule out that some sparse architectures may be more stable than others. 
Recurrent neural nets (RNNs), however, have underlying geometric properties which may allow for greater interpretability of sparse architectures: Sussillo and Barak (2013), for instance, show that the hidden state of various low-dimensional RNNs is concentrated along \textit{slow manifolds} of a dynamical system \cite{sussillo2013openingtheblackbox}, also described as \textit{attractor manifolds}. 

In this paper, we provide evidence that the geometry of RNN hidden states can be parlayed into geometric structures underlying sparse RNNs. We do this by combining magnitude pruning, in which the lowest valued weights of the neural net are thresholded during training, with \emph{moduli regularization}. Moduli regularization, which we introduce in Section \ref{sec3:moduli}, is a topologically inspired regularization term applied to the hidden-state update matrix of the RNN. It is computed by embedding hidden-state neurons into a metric space $\mathcal{M}$, and differentially penalizing weights according to their distance on $\mathcal{M}$. We further show that the geometry of $\mathcal{M}$ can be implicitly learned and implemented during training at relatively low cost. 

Our approach reflects the continuous attractor theory of RNNs \cite{seung1997learning}. The continuous attractor theory posits that the hidden layer of an RNN isolates a low dimensional parameter space (the eponymous continuous attractor, or the moduli space) of stable states, and updates at each time correspond to small movements on this space. From this perspective, moduli regularization is a method of training RNNs which favors particular embeddings of the continuous attractor into the hidden state space: specifically, embeddings which are sparse in the neural basis. By specifying neural connections based on distance on a manifold, we perform the inverse to a common strategy for manifold detection in neuroscience, in which a manifold is discovered by linking neurons that simultaneously fire \cite{gardner2022toroidal}.



We test the efficacy of our method in three recurrent architectures: a navigation RNN \cite{cueva2018emergence}, a natural language processing RNN \cite{pytorch_examples}, and an adding problem RNN \cite{le2015simple}.
In each case, we show that a moduli space well-suited to the RNN creates superior sparse models as opposed to comparable methods. We then randomize the model's weights and retrain the neural net on the sparse architecture to test the stability of the learned sparse architecture, showing that moduli regularization not only provides superior sparse models, but also produces sparse architectures which are more stable than alternative methods. 

To the best of our knowledge, this work is the first to demonstrate the role of structured \textit{a priori} sparsity in RNNs. It further provides new evidence that global topological structures, rather than purely local network properties, play a critical role in determining the performance and stability of sparse neural architectures.

The key contributions of this work are:

\begin{itemize}
\item We introduce \textit{moduli regularization} as a framework for highly efficiently sparsifying RNNs using geometric principles. 
\item We introduce the first method for end-to-end explicit learning of underlying geometric structures during RNN training. 
\item  We provide empirical validation of our method, highlighting the importance of global topological structure, as opposed to purely local network structures, in sparse network design.
\end{itemize}

Our approach advances a structured and principled methodology for sparsity in neural networks, addressing both theoretical and practical aspects of efficient and robust model design.

\subsection{Related work}
\noindent {\bf Continuous attractors.} Attractor dynamics in RNNs and their utility in image classification were first proposed by \cite{seung1997learning}. Many recent papers discover continuous attractors by analyzing the hidden state of trained RNNs \cite{elman1991distributed, schaeffer2023self, sorscher2023unified, sussillo2013openingtheblackbox}. Similar continuous attractors, and handmade networks that instantiate them, are common in the analysis of neural firing in the brain \cite{burak2009accurate, gardner2022toroidal, guanella2007model}. 
Work to encode favorable properties in RNNs, such as translation equivariance \cite{zhang2022translation}, are also studied in hand-crafted neural nets. However, these studies lack methods to inform the training of general RNNs with their underlying attractors.

\noindent {\bf Topology. } The incorporation of parameterizing manifolds into neural network architectures has received some attention: architectures built around Grassmannians \cite{DBLP:journals/corr/HuangWG16}, symmetric positive-definite matrices \cite{DBLP:journals/corr/Huang0LLSGC16, konstantinidis2023multimanifold}, and the Klein bottle \cite{topdeeplearning} have been proposed for various tasks. These approaches are built around \emph{a priori} knowledge of data moduli spaces, which is then built into the architecture. Other authors have attempted to combine simplicial complexes learned from the data \cite{osti_10303167} and from neural data recorded from mice \cite{mitchell2024topological}. 

\noindent {\bf Manifold discovery. } A foundational challenge for topologically aware neural nets is manifold discovery, or the learning of low dimensional manifolds that well describe given data \cite{fan2021manifold, tenenbaum2000global}. There is growing interest in the topological data and machine learning communities regarding the use of mixed curvature representations as a replacement for standard Euclidean representations of both data and learned features. This allows representations that flexibly match the true curvature of the dataset by learning both optimal curvature values for each product component \cite{gu2019learning, skopek2020mixed} and the distribution of curvature signs throughout the product \cite{saez2024neural}. These techniques find notable improvements over baseline representations in variational autoencoders \cite{saez2024neural}, embedding spaces \cite{gu2019learning}, and learned representations for downstream tasks \cite{saez2024neural}.

\noindent {\bf Sparsity. } Investigations of regularization in recent literature includes both methods focused on achievement of fixed sparsity targets \cite{NEURIPS2022_sparsity, NEURIPS2022_089b592c_controlledsparsity, louizos2017learning, singh2020woodfisher} and methods which sparsify differently depending on the distribution of weights \cite{narang2017exploring}. However, in large models, simple magnitude pruning remains the state of the art in sparsification \cite{gale2019state}. Training \emph{ab initio} sparse models remains a fundamental challenge: sparse models are less stable than dense ones, meaning that random weight initializations are less likely to converge to high fidelity models \cite{frankle2018lottery}. Even with identical weight initializations, Frankle et al. \cite{frankle2020linear} show that the stability of network training with respect to randomness in stochastic gradient descent was achieved only after a degree of training, at which point the corresponding sparsified networks agreed.

The Lottery Ticket Hypothesis of \cite{frankle2018lottery} is a proposed explanation for the paradox of the effectiveness of large-scale sparsification, and the ineffectiveness of training \emph{a priori} sparse models \cite{gale2019state}. It argues that the large over-parameterization of neural nets makes it very likely that some subnetwork is configured with a fortunate set of weights. 
Much past work on sparse neural nets focuses on convolutional layers of image recognition neural networks, where benchmarks are well established \cite{gale2019state, louizos2017learning, molchanov2017variational, zhu2017prune}, though attention has also been given to transformers \cite{gale2019state, vaswani2017attention} and recurrent neural nets \cite{narang2017exploring} for natural language processing.

\section{Continuous attractors}

\subsection{Preliminaries}
A recurrent neural net is a general term for any neural net designed to analyze sequences or streams of data, one element at a time. The Elman Recurrent Neural Network, henceforward abbreviated as RNN, is one of the most straightforward yet effective structures in common usage \cite{elman1991distributed}. The RNN receives a stream of data $\{x_0, x_1, x_2, ...\}$ represented as vectors, and maintains a hidden state $H_t$ at each time $t$, a vector that functions as an internal representation of the state of the system. After an initial hidden state $H_{-1}$ is fixed, the $H_t$ are computed recursively as
\begin{equation} \label{eq:elman_hidden_update}
H_t = s(W_{hh}H_{t-1} + W_{ih}x_t + B), 
\end{equation}
for $t \ge 0$, where $W_{hh}, W_{ih}$ are appropriately dimensioned matrices, called the hidden update and input matrices, respectively, $B$ is the bias vector, and $s$ is a nonlinear function, typically $\tanh$, or $\on{ReLu}$. This is represented in Figure \ref{fig3:mod_explanation}(a). A decoding matrix $D$ then takes the final hidden state, $H_t,$ of an RNN to the desired output,
\begin{equation} \label{eq:elman_decoder}
    \text{output} = s'(D(H_t) + B_d),
\end{equation}
where $B_d$ is a vector, called the decoder bias, and $s'$ is an activation function, such as the softmax (Equation \eqref{eq:softmax}). A multi-layer RNN is a stack of recurrent neural networks: the first recurrent layer takes as input the input vectors $x_t$, as in Equation \eqref{eq:elman_hidden_update}. Subsequent recurrent layers take as input the previous layer's hidden state $H_t$. For an $n$-layer RNN, with fixed initial hidden states $\{H_{-1, k}\}_{k=0,...,n-1}$, the hidden states are updated recursively as
\begin{equation}\label{eq:multi_layer_rnn}
    H_{t,k} = s(W_{hh}^kH_{t-1,k} + W_{ih}H_{t,k-1} + B),
\end{equation}
where $s, W, H, B$ are as above when $k > 0$, and using the update of \eqref{eq:elman_hidden_update} when $k = 0$.

An RNN can be viewed as a discrete approximation of a dynamical system on $\BR^n$. The RNN acts as a continuous automorphism of $\BR^n$, which induces a vector field on $\BR^n$. A vector field is an assignment, for each point $x \in \BR^n$, of a vector $v(x) \in \BR^n$ such that $v(x)$ is smoothly varying in $x$. 
To flow along a vector field from a point $x_0$ means to take small steps in the direction $v(x_t)$ along a sequence of points $x_t = x_{t-1} + v(x_{t-1})dt$. A fixed point $x$ of a vector field is a point with velocity $v(x) = 0$. In addition, such a point $x$ is called stable if it flows from points in a neighborhood $U$ converge to $x \in U$. A continuous attractor associated to a vector field $\CA$ is a continuous locus of stable points. 

Continuous attractors have a \emph{metric structure}, which means they are equipped with a distance function which satisfies the triangle inequality, is symmetric, and satisfies $d(x, y) = 0$ if and only if $x=y$. A smooth manifold is a special kind of topological space which is locally homeomorphic to $\BR^n$ (see Appendix \ref{appendix:manifolds} for details), and on which derivatives can be defined. Smooth manifolds are metric spaces whose distance function is called geodesic distance. Geodesic distance measures the length of the shortest path between two given points. Common examples of manifolds include a circle, sphere, torus, and $\BR^n$.

\subsection{Moduli regularization for RNNs}\label{sec3:moduli}
A common technique to help induce model sparsity is $L_1$ regularization. $L_1$ regularization is applied to a model $M: \mathcal{X} \to \mathcal{Y}$ with criterion $\on{criterion}(M(x), y)$ by minimizing the following loss function:
\begin{equation}\label{eq:regularized_loss_func_l1}
\on{loss}(M, x, y) = \on{criterion}(M(x), y) + \lambda \sum_{s \in S} |w_s|.
\end{equation}
Here $\{w_s\}_{s \in S}$ is a list of all weights in the neural net and $\lambda$ is the regularizing constant. More generally, we may vary the regularizing coefficient for each weight or subset of weights, as in the following sample loss function:
\begin{equation}\label{eq:regularized_loss_func_general}
\on{loss}(M, x, y) = \on{criterion}(M(x), y) + \lambda \sum_{s \in S} c_s|w_s|.
\end{equation}

$L_1$ regularization \cite{tibshirani96,l1_paper} does not directly create sparse networks, and much work \cite{gale2019state, louizos2017learning, molchanov2017variational, narang2017exploring, zhu2017prune} has gone into creating different regularization techniques that will more directly induce sparsity. Our method, which we term \emph{moduli regularization}, is rooted in the hypothesis that manifolds (Appendix \ref{appendix:manifolds}) usefully approximate the hidden space of RNNs. This is a form of the manifold hypothesis, which suggests that high dimensional data is often encoded in a low dimensional manifold \cite{fefferman2016testing}. The manifold hypothesis has led to a wide variety of techniques for manifold learning and machine learning (e.g. \cite{tenenbaum2000global, fan2021manifold, saez2024neural, skopek2020mixed}). In RNNs the manifold hypothesis has been studied primarily in terms of manifold emergence during learning, such as the slow manifolds of \cite{sussillo2013openingtheblackbox} for state-change tasks, or toroidal grid cell emergence \cite{cueva2018emergence, gardner2022toroidal, sorscher2023unified, schaeffer2023self}. Moduli regularization proposes immediate exploitation of the manifold structure to create superior sparse RNNs.

\begin{figure}
    \centering
    \includegraphics[width=.75\linewidth]{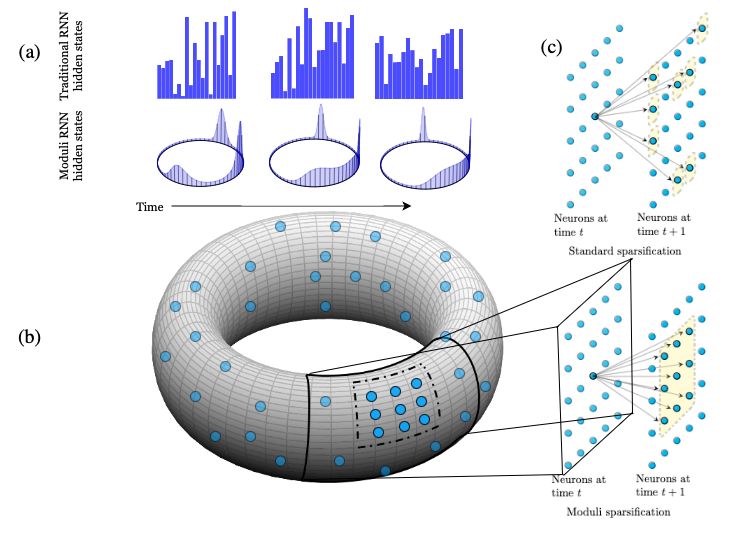}
    \caption{(a) Top: a traditional RNN tracks hidden states as a sequence of vectors in $\BR^n$, without fixed structure; each bar represents the number stored at that dimension in the RNN. Bottom: with moduli regularization, the hidden state approximates a function on the circle, by using neurons (marked by dark lines) as a discrete approximation of the circle. These neurons record values approximating a distribution, represented by the height of the dark lines. (b) Hidden state neurons, represented by blue dots, are embedded into a moduli space, the torus. (c) Sparsification of the hidden update matrix of an RNN (Equations \ref{eq:elman_hidden_update}, \ref{eq:multi_layer_rnn}). Above depicts random sparsification, and below depicts sparsification in line with moduli regularization (briefly, \textit{moduli sparsification}). Yellow boxed points are neurons with a non-zero weight connecting them to the center neuron. Moduli sparsification respects the geometry of the chosen moduli space, which is ignored by standard sparsification techniques.}
    \label{fig3:mod_explanation} 
\end{figure}

Fix a metric space $(\mathcal{M}, d_{\mathcal{M}})$. Let $\{e_1, ..., e_n \}$ be the standard basis vectors for the hidden state $\BR^n$ of an RNN, and choose an embedding $i: \{1, ..., n\} \to \mathcal{M}$ from the set of hidden state neurons into $\mathcal{M}$. Let $W_{hh} = (w_{jk})_{1 \le j, k \le n}$ be the hidden update matrix of Equation \eqref{eq:elman_hidden_update}. For a function $f: \BR_{\ge 0} \to \BR_{\ge 0}$ and $\ell \ge 1$, define an $f$-regularizer $R_f: \BR^{n^2} \to \BR$ of $W_{hh}$ associated to the embedding $i$ to be
\begin{equation} \label{eq:moduli_reg}
R_f(W_{hh}) := \sum_{j,k} f(d_{\mathcal{M}}(i(j), i(k)))|w_{jk}|^\ell. 
\end{equation}
We depict the structure of the regularizer in Figure \ref{fig3:mod_explanation}. In the figure, $f$ is monotonically increasing for visual clarity, though this need not be true in practice (see Appendix \ref{appendix:inhibitory_functions}). 

The function $f: \BR_{\ge 0} \to \BR_{\ge 0}$ is referred to as the \emph{inhibitor function}. The inhibitor function is an additional parameter of the regularizer, which dictates the manner in which information is passed along the metric space. For example, explicit constructions of continuous attractors in neural nets for navigation \cite{burak2009accurate, guanella2007model} are constructed with connectivity given by the Ricker wavelet
\begin{equation}\label{eq:ricker}
    \psi(d) = \frac{2}{\sqrt{3\sigma\sqrt{\pi}}}\left(1 - \frac{d^2}{\sigma^2}\right) e^{-\frac{d^2}{2\sigma^2}},
\end{equation}
where $d$ is the distance between two points on the moduli space.

We regularize each layer of multi-layer RNNs independently on the same manifold by choosing three different embeddings and using these to generate regularizers. It is also possible to change the manifold between recurrent layers, however we do not present any such experiments. Furthermore, this framework could be studied for non-recurrent architectures, although that is not the focus of the current work (see Appendix \ref{appendix:non_recurrent_layers}).


\subsection{Learning the moduli space}\label{sec3:trained_embeddings}
The moduli regularizer involves the choice of an embedding $i: \{1, 2, ..., n\} \to \mathcal{M}$ in addition to the choice of $\mathcal{M}$. We initialize our embeddings by drawing points independently and uniformly at random over $\mathcal{M}$ when $\mathcal{M}$ is compact. When $\mathcal{M}$ has a smooth structure and the inhibitor function $f$ is differentiable, $i$ can be taken as a parameter of the RNN and optimized by gradient descent during the course of training. By choosing $\mathcal{M}$ to be Euclidean space $\BR^k$, the embeddings can cluster on a submanifold, explicitly learning desirable geometric relationships---moreover, any manifold of dimensions less than or equal to $k/2$ can be perfectly approximated by this process, by the Whitney embedding theorem. 


Monotonically increasing inhibitor functions cannot be effectively combined with trained embeddings: if used, the moduli regularizer (Equation \eqref{eq:moduli_reg}) is minimized by any embedding $\{j\}_{1 \le j \le n} \to * \hookrightarrow \mathcal{M}$ which maps all neurons to a single point in the moduli space. Moduli regularization then degenerates to $L_{\ell}$ regularization with regularizing coefficient $f(0)$. Other choices of $f$, however, such as the difference of Gaussians
\begin{equation}\label{eq:DoG}
    f(x) = c - c(e^{-\frac{x^2}{\sigma_2}} - e^{-\frac{x^2}{\sigma_1}}),
\end{equation}
where $c > 0, \sigma_2 > \sigma_1 > 0,$ or the Ricker wavelet (Equation \eqref{eq:ricker}) will not degenerate, and allow effective training of the embedding.

Learning the moduli regularizers comes at the detriment of adding $n \cdot \on{dim} \mathcal{M}$ trained parameters to the model. However, $\on{dim} \mathcal{M}$ is in practice substantially smaller than $n$, hence the additional trained data is a small fraction of the trained data of the RNN. Additional overhead computations are also necessary for the recalculation of the regularizing terms, requiring $O(n^2)$ computations at the end of gradient step epoch. In practice, we found this overhead was negligible when $n < 500$, and small when $n < 1000$. 

\section{Limitations}\label{sec3:limitations}
Moduli regularization shows some sensitivity to the regularizing factor, a common limitation for regularization techniques. The choice of moduli space $\mathcal{M}$ and inhibitor $f$ are further additional hyperparameter which can be hard to properly select, and are slow to be learned when working with RNNs with very high dimensional hidden spaces. Interesting avenues for future research include choosing a moduli space using manifold discovery, a wide range of techniques for storing data and computing optimal, nonlinear representation schemes \cite{fan2021manifold, gu2019learning, DBLP:journals/corr/HuangWG16, saez2024neural, skopek2020mixed}.

One possible conflating factor for our experiments is our sparsification scheme---magnitude pruning. Magnitude pruning is the simplest pruning method available, and we chose it in order to simplify the discussion and avoid conflating factors in our experiments. Changes to this sparsification method might impact the level of improvement moduli regularization offers, as compared to alternative regularization schemes, though our lottery ticket hypothesis \cite{frankle2018lottery} experiments (Section \ref{sec3:experiments}) will remain of interest in this comparison. Finally, due to the computational costs, we did not extensively optimize the hyperparameters in our experiments, and tested on limited architectures.

\section{Results}\label{sec3:experiments}
We test moduli regularization (Equation \eqref{eq:moduli_reg}) on three benchmark tasks: navigation, natural language processing (NLP), and the adding problem.
Navigation is a geometric regression task, with known moduli spaces on which the recurrent dynamics occur in both mammalian brains \cite{burak2009accurate, gardner2022toroidal, guanella2007model} and simulations \cite{cueva2018emergence, schaeffer2023self, sorscher2023unified}. NLP
is a classification task which lacks clean geometric descriptions \cite{elman1991distributed}. The adding problem is an experiment designed to study the ability for RNNs to maintain long term memories, and likewise lacks obvious geometry.  Despite this distinction, we show that moduli regularization is a highly effective technique for model sparsification for all benchmarks. 

We study these tasks using both trained and untrained moduli regularizers in a variety of low dimensional ambient spaces $\mathcal{M}$ for comparison. We find that training the regularizer, assuming lack of knowledge of a good moduli space, is highly effective for both navigation and adding problem experiments. In our NLP experiments, however, training the regularizer seemed to enable additional overfitting to occur, as opposed to fixing random embeddings. 


We use magnitude pruning to sparsify models during training. This means that to train models which are $p$\% sparse over $n$ training epochs, we threshold the lowest magnitude $(k-1)p/(n-1)\%$ of the model's weights to 0 at the beginning of the $k^{\text{th}}$ epoch. In our experiments, we chose $p$ by running initial tests and hand selecting values at which sparse model quality was beginning to break down.

The lottery ticket hypothesis \cite{frankle2018lottery} suggests that sparse architectures arise from fortunate weight initializations. In other words, the training process for sparse neural architectures is unstable: the subspace of potential weight initializations which will cause training a model to converge to a high-quality net is low measure in the space of all potential weight initializations. This contrasts with dense neural nets, which are robust to random weight initializations. As a result, retraining models on a sparse architecture without using the original weight initialization deteriorates performance \cite{frankle2018lottery, gale2019state}. We show that the sparse structures learned with moduli regularization are more stable than the sparse structures learned without it: models retrained on moduli-generated sparse architectures outperform models retrained on traditional sparse architectures. In several of our NLP experiments, we even achieve superior results after reinitialization and retraining \emph{ab initio} sparse models as opposed to the original sparsified models. For all experiments, we report both the error of progressively sparsified models, labeled $X$\% Sparse, and the results of retraining with randomized weights in the learned sparse architecture, labeled $X$\% Lottery. Lottery ticket experiments were additionally conducted with the same regularizer as was used in the antecedent model. When using trained embeddings, we ceased training embeddings for lottery ticket experiments (Tables 1--9).


\subsection{Navigation}\label{sec3:navigation}
We study a single-layer RNN designed for autonomous landmark mapping and navigation, using the architecture of \cite{cueva2018emergence}. A 2.2 m square arena is fixed, and $\ell$ uniformly randomly chosen points in the arena are fixed throughout training, and referred to as landmarks. The RNN initializes its hidden state with a two-dimensional position $x_0$, which is transformed to $\on{softmax}(d(x_0, p_i))_{i=1,...,\ell}$, where $d$ computes distance in $\BR^2$. The recurrent input to the RNN is a sequence of two-dimensional velocity vectors $\{v_1, v_2, ..., v_{seq\textunderscore length} \}$, which assemble into a trajectory in the arena. The final hidden state of the RNN is mapped linearly back to $\BR^\ell$, and softmax is applied. The loss term is the cross entropy of the output with $\on{softmax}(d(x_{seq\textunderscore length}, p_i))_{i=1,...,\ell}$, where $x_{seq\textunderscore length} = x_0 + \sum_i v_i$ is the final position of the trajectory. We illustrate this set-up in Figure \ref{fig:navigation_explainer}, and in greater detail in Supplemental Figure\ref{fig:appendix_navigation_explainer}. We then applied a Bayesian decoder to infer most likely position in the arena, based on the activations in $\BR^\ell$, and report the distance between the predicted and true position. Untrained models have a mean error of $\approx 100$ cm in arenas of this size. Training data in the form of trajectories can be freely generated for the RNN, which alleviates the problem of overfitting. This allows us to study the effect of regularization on the sparsity of the underlying model, isolated from the other benefits of regularization. 
Hyperparameters for our experiments are listed in Appendix \ref{appendix:rnn_hyperparameters}.  

\begin{figure}
    \centering
    \includegraphics[width=0.6\linewidth]{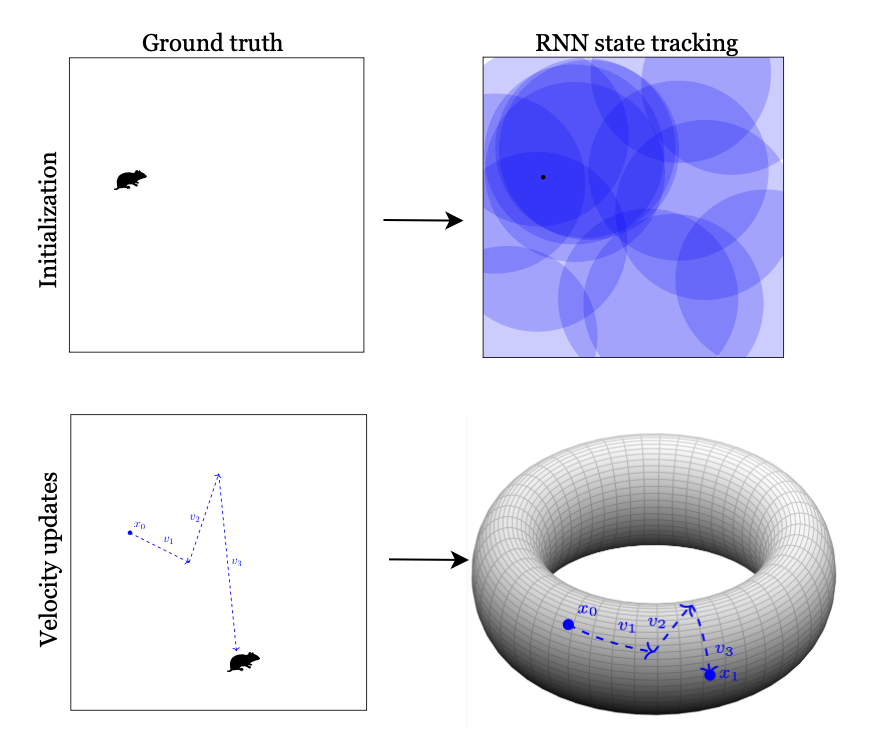}
    \caption{The representation of the state transitions in the RNN. Left: ground truth of the navigation problem. Right: The internal representation used by a (toroidally regularized) RNN. Top: initialization in the ground truth places the agent at a random place in a box. The RNN uses a Gaussian tiling to represent the initial position. Bottom: In the ground truth, a sequence of velocity inputs moves the agent to a new position. In the RNN, this is represented by changing the hidden state to reflect the new position. In the figure, this is depicted by translation on a torus. }
    \label{fig:navigation_explainer}
\end{figure}



{\bf Control experiments:} To evaluate the utility of moduli regularizers, we compare toroidal $L_1$ regularization to models without regularization, with standard $L_1$ regularization, and with shuffled moduli regularizers. 
To construct shuffled regularizers, we generate moduli regularizers then randomly permute the regularizing coefficients (Equation \eqref{eq:regularized_loss_func_general}).
Table \ref{tab3:nav_control} shows that while shuffled moduli regularizers perform well, they perform worse than moduli regularization: the inclusion of topology in the regularizing term greatly improves the sparse model fidelity. 
Both shuffled moduli and moduli sparsification are superior to $L_1$ regularization and unmodified magnitude pruning, two of the most commonly used techniques for reducing model complexity.

\begin{table}[]
    \centering
    \caption{Mean navigation error in cm (standard deviation) over five experiments. Only toroidal moduli regularization accommodates 90\% sparsification of the model, either during the original training or while testing the lottery ticket hypothesis.
    }
\begin{tabular}{lcccc} \toprule
 & Torus & No reg. & $L_1$ & Shuffled  \\ \cmidrule{2-5}
 0\% Sparse Err. & 4.67 (0.07) & \textbf{4.61} (0.04) & 5.04 (0.03) & 5.18 (0.07)  \\
75\% Sparse Err. & \textbf{4.86} (0.30) & 39.03 (41.95) & 14.73 (12.23) & 5.37 (0.07) \\
75\% Lottery Err. & \textbf{7.28} (0.69) & 27.42 (33.28) & 14.17 (2.18) & 7.29 (0.81) \\
90\% Sparse Err. & \textbf{5.69} (0.43) & 71.75 (32.67) & 86.70 (5.95) & 10.92 (0.87)  \\
90\% Lottery Err. & \textbf{12.49} (2.03) & 64.44 (3.81) & 53.59 (9.39) & 42.25 (7.65)  \\ \bottomrule
\end{tabular}
    \label{tab3:nav_control}
\end{table}


\noindent {\bf Trained embeddings:} In Table \ref{tab:trained_r3}, we compare trained embeddings (as in Section \ref{sec3:trained_embeddings}) to fixed embeddings for both the torus and $\BR^3$. $\BR^3$ is a space poorly suited towards navigation but contains apt moduli spaces such as the torus as a subspace. On the torus, this fails to materially improve the model. However, when  $\BR^3$ is used, training embeddings results in substantial improvements to the model. We additionally apply persistence homology to the trained neurons to attempt to discover what subspace of $\BR^n$ the neurons gravitate towards in Appendix \ref{appendix:trained_persistence}. 
\begin{table}[t]
    \centering
    \caption{Mean navigation error in cm (standard deviation) experiments of trained embeddings in the torus and $\BR^3$ over five and three experiments, respectively. Trained embeddings have little effect in an apt moduli space, such as the torus, but result in substantial improvements as opposed to randomly chosen embeddings in $\BR^3$.}
\begin{tabular}{lcccc} \toprule
 & Torus trained & Torus fixed & $\BR^3$ trained  & $\BR^3$ fixed \\ \cmidrule(l{2pt}r{4pt}){2-3} \cmidrule(l{4pt}r{2pt}){4-5}
0\% Sparse & 4.79 (0.10) & \textbf{4.67} (0.07) & 5.85 (0.09) & \textbf{4.77} (0.09) \\
75\% Sparse  & \textbf{4.60} (0.06) & 4.86 (0.30) & \textbf{6.59} (0.85) & 34.78 (42.10)\\
75\% Lottery  & 8.20 (4.16) & \textbf{7.28} (0.69) & \textbf{35.44} (39.83) & 60.37 (35.54) \\
90\% Sparse  & 5.87 (0.55) & \textbf{5.69} (0.43) & \textbf{5.81} (0.45) & 67.42 (33.27) \\
90\% Lottery  & 15.72 (3.94) & \textbf{12.49} (2.03) & \textbf{20.67} (5.68) & 46.56 (7.69) \\
\bottomrule
\end{tabular}
    \label{tab:trained_r3}
\end{table}

{\bf Moduli spaces:} Table \ref{tab3:nav_moduli_comparison} compares the use of different moduli spaces in the regularizer. The torus and the Klein bottle both provide superior sparse models, comporting with their admission of $\BR^2$ as a universal cover (see Appendix \ref{appendix:covering_spaces}). The circle also performs well, likely because $S^1 \times S^1$ is isomorphic to the torus.

\begin{table}[]
    \centering
    \caption{Mean navigation error in cm (standard deviation) over five experiments. As a moduli regularizer, the torus, the circle, and the Klein bottle accommodate 90\% sparsification, both in original training and when testing the lottery ticket hypothesis. }
\begin{tabular}{lccccc} \toprule
 & Torus & Circle & Sphere & Klein & 6 Torus \\ \cmidrule{2-6}
 0\% Sparse & \textbf{4.70} (0.07) & 14.47 (12.98) & 35.09 (20.73) & 4.73 (0.08) & 8.29 (1.66) \\ 
 75\% Sparse & 4.86 (0.30) & 4.80 (0.25) & 22.93 (34.16) & \textbf{4.79} (0.12) & 21.02 (13.25) \\
 75\% Lottery & 7.28 (0.69) & 6.93 (0.60) & \textbf{6.45} (0.71) & 6.56 (0.40) & 66.75 (10.89) \\
 90\% Sparse & 5.69 (0.43) & 5.72 (0.71) & 23.55 (35.05) & \textbf{5.44} (0.13) & 34.71 (16.60) \\
 90\% Lottery & 12.49 (2.03) & 11.02 (1.49) & 28.33 (11.85) & \textbf{10.96} (1.71) & 93.93 (0.73) \\ \bottomrule
\end{tabular}
    \label{tab3:nav_moduli_comparison}
\end{table}



Sensitivity analysis on the regularizing factor ($\lambda$ in Equation \eqref{eq:regularized_loss_func_general}) can be found in Appendix \ref{appendix:ablation}. 
This RNN architecture was originally introduced by \cite{sorscher2023unified} to study grid cells, certain specialized neurons in the entorhinal cortex which provide an internal map of space \cite{giocomo2011computational, moser2008place, whittington2022build}. After the discovery of grid cells in 2005 \cite{hafting2005microstructure}, several authors \cite{burak2009accurate, guanella2007model} show that toroidal neural connectivity can give rise to grid cell structures. However, to the best of our knowledge, no connectivity structures outside of the torus and $\BR^2$ have been seriously investigated---the hand construction of these networks is time consuming, and small perturbations of weights can damage their effectiveness \cite{vafidis2022learning}. One surprising application of our moduli regularization is then the great ease with which it shows the learning advantages of the torus over arbitrary spaces, such as the sphere, for neural attractor dynamics on navigation. The success of toroidal, circular, and Klein bottle sparsification suggests that moduli regularization may provide a useful lens for exploratory work on highly structured neural connectivity. We expected these regularizers to outperform the sphere and 6-torus because $\BR^2$ is a covering space (see summary in Appendix \ref{appendix:covering_spaces}) for the Klein bottle and the torus, making them apt compact representations for two-dimensional space. Since the torus is isomorphic to the product of two circles, a circular moduli space could also plausibly recreate toroidal connectivity. 


\subsection{Natural language processing}\label{sec3:nlp}
Natural language processing lacks clear \emph{a priori} geometry to exploit \cite{elman1991distributed}. Indeed, local stability of the hidden state is not a common hypothesis for NLP architectures like long-short term memory \cite{hochreiter1997long}. We test an RNN with three recurrent layers trained to predict subsequent words in a sequence on the Wikitext-2 dataset of high-quality Wikipedia articles \cite{merity2016pointer}. Our implementation is based on the implementation of \cite{pytorch_examples}, which is a representative implementation of natural language RNNs. Hyperparameters for our experiments are listed in Appendix \ref{appendix:word_architecture}. 

\noindent {\bf Control experiments:} To evaluate the importance of the underlying topology of moduli regularization, we compare toroidal $L_1$ regularization with difference of Gaussians inhibitor (Equation \eqref{eq:DoG}) to models trained without regularization, with $L_1$ regularization (Equation \eqref{eq:regularized_loss_func_l1}), and with shuffled regularization values (Table \ref{tab3:control_word}). Here as above, `Shuffled' means that we construct a toroidal moduli regularizer, but randomly permute the coefficients of the regularizing term. At baseline 0\% sparsity, toroidal moduli regularizers perform slightly worse than their shuffled version, reflecting the positive effects of random sparsification in non-geometric problems, and the importance of finding the true moduli space for NLP applications. Nevertheless, when incorporating sparsification, we find that the toroidal geometry provide some advantages compared to random sparsification. Both toroidal and shuffled moduli regularizers provide substantial improvements compared to $L_1$-regularizers and non-regularized systems.

\begin{table}[]
    \centering
    \caption{Mean test perplexity (standard deviation) of next-word prediction RNNs at 0\%, 90\%, and 98\% sparsity, and testing the lottery ticket hypothesis with trained sparse architectures. $\,^*$ One outlier experiment with the $L_1$ regularizer, marked by a star, suffered representation collapse and had final perplexity $\approx 3.4 \times 10^{10}$. We excluded this experiment from reported averages.}
\begin{tabular}{lccccc} \toprule
 & Torus & No reg. & $L_1$ & Shuffled  \\ \cmidrule{2-5}
0\% Sparse & 69.26 (16.31) & 114.73 (1.56) & 172.19 (94.41) & \textbf{56.52} (5.64) \\ 
90\% Sparse & \textbf{43.41} (7.53) & 91.12 (34.21) & 118.18 (4.03) & 67.98 (34.00) \\
90\% Lottery & \textbf{54.80} (8.70) & 113.93 (0.92) & 114.41 (0.92) & 75.17 (21.55) \\
98\% Sparse & 55.80 (13.96) & 118.45 (1.59) & $98.78 (31.97)^*$ & \textbf{51.06} (10.15) \\
98\% Lottery & \textbf{61.44} (6.06) & 115.21 (1.27) & 115.06 (1.88) & 62.58 (5.57) \\\bottomrule
\end{tabular}
    \label{tab3:control_word}
\end{table}

\noindent {\bf Trained embeddings:} As in navigation, we studied trained embeddings on the torus and in $\BR^3$. To our surprise, training the embeddings resulted in very similar results to the navigation experiment, showing improvements in $\BR^3$, but performing poorly on the torus. We give results of our experiments in Table \ref{tab:nlp_trained_embeds}. We suspect this deficiency might reflect the regularizer abetting overfitting in the trained case, though we did not extensively investigate this hypothesis. 

\begin{table}[]
    \centering
        \caption{Mean perplexity (standard deviation) of five experiments at 0\%, 90\%, and 98\% sparsity in a next-word-predictor RNN. Training embeddings as we did in this context can be actively harmful to the final model quality. }
\begin{tabular}{lcccc} \toprule
 & Torus trained & Torus fixed & $\BR^3$ trained & $\BR^3$ fixed  \\ \cmidrule(l{2pt}r{4pt}){2-3} \cmidrule(l{4pt}r{2pt}){4-5}
 0\% Sparse & 119.90 (5.92) & \textbf{69.26} (16.31) & \textbf{109.49} (19.83) & 116.18 (1.97)  \\
90\% Sparse & 45.70 (7.29) & \textbf{40.35} (4.30)& \textbf{68.53} (19.21) & 117.25 (1.54)  \\
90\% Lottery & 109.56 (5.96) & \textbf{56.43} (2.71) & 114.04 (0.37) & 114.56 (0.00)  \\
98\% Sparse & 107.24 (24.92) & \textbf{65.53} (9.40) & \textbf{62.52} (10.46) &   121.37 (4.02) \\
98\% Lottery & 115.36 (0.65) & \textbf{62.57} (7.54) & \textbf{107.92} (10.52) &  116.37 (0.67) \\
\end{tabular}
    \label{tab:nlp_trained_embeds}
\end{table}

\noindent {\bf Moduli spaces:} We next study an array of moduli regularizers using different moduli spaces, shown in Table \ref{tab3:word_moduli}. All experiments use the difference of Gaussians inhibitor function (Equation \eqref{eq:DoG}). Unlike in navigation, our experiments show no clear preference towards specific moduli spaces, suggesting that none of our experiments reflect a true moduli space for NLP. At 90\% sparsity, all moduli regularizers outperformed all control experiments, but this benefit degrades at 98\% sparsity. 

\begin{table}[]
    \centering
    \caption{Mean test perplexity (standard deviation) of next-word prediction RNNs at baseline, 90\%, and 98\% sparsity, trained with a variety of moduli regularizers, and testing the lottery ticket hypothesis. Underlined values are \emph{ab initio} sparse models that outperform their sparsified antecedents.}
\begin{tabular}{lccccc} \toprule
 & Torus & Circle & Sphere & Klein & 6-Torus \\ \cmidrule{2-6}
0\% Sparse & 69.26 (16.31) & 58.03 (15.78) & 65.51 (6.48) & 64.35 (10.89) & 57.17 (12.80) \\
90\% Sparse & \textbf{43.41} (7.53) & 49.33 (10.61) & 49.38 (10.76) & 51.03 (10.46) & 44.91 (3.63) \\
90\% Lottery & 54.80 (8.70) & 51.32 (11.08) & 55.52 (11.56) & 57.78 (10.87) & \textbf{46.16} (6.70) \\
98\% Sparse & 55.80 (13.96) & \textbf{44.88} (3.32) & 52.86 (12.57) & 58.86 (7.44) & 57.68 (10.61) \\
98\% Lottery & 61.44 (6.06) & 63.84 (3.69) & 57.89 (5.79) & \textbf{53.34} (2.18) & 63.73 (5.78) \\ \bottomrule
\end{tabular}
    \label{tab3:word_moduli}
\end{table}
\begin{table}
\centering 
\caption{Mean test perplexity (standard deviation) on RNNs with trained embeddings. Trained embeddings in the torus allows the model to overfit, while it improves the  moduli approximation in $\BR^3$. }
\begin{tabular}{ccccc} \toprule
 & R3 trained & R3 fixed & Torus trained & Torus fixed \\ \cmidrule(l{2pt}r{4pt}){2-3} \cmidrule(l{4pt}r{2pt}){4-5}
90\% Sparse & 81.50 (27.70) & 106.38 (21.23) & 103.22 (27.56) & 43.41 (7.53) \\
90\% Lottery & 101.93 (15.26) & 114.25 (0.39) & 108.59 (12.42) & 54.80 (8.70) \\ \bottomrule
\end{tabular}
\label{tab3:NLP_trained_embeddings}
\end{table}


Our results show that moduli regularization remains a highly effective sparsification method in non-geometric RNNs. Moduli regularizers  (Table \ref{tab3:word_moduli}) improve the accuracy of trained NLP models compared to both non-regularized and $L_1$ regularized models (Table \ref{tab3:control_word}).  While they do not consistently outperform random sparsification at 98\% sparsity, some moduli regularizers create sparse architectures that are more resilient to retraining with randomized weights, in some cases defying the lottery ticket hypothesis \cite{frankle2018lottery}.  

\subsection{The adding problem}\label{sec:adding_problem_experiments}
The adding problem is an experiment to study long term memory in recurrent architectures (see \cite{le2015simple}, Section 4.1, or \cite{li2018independently}, Section 5.1). The input data is a sequence of numbers chosen uniformly at random between 0 and 1, as well as a mask which specifies 1 at two points in the sequence, and 0 elsewhere. The target for the RNN is the sum of the product of the input data row with the mask row. We illustrate an example of the adding problem in Figure \ref{fig:adding_problem}. Hyperparameters for our experiments are listed in Appendix \ref{appendix:adding_problem_hypers}, and all inhibitor functions are the difference of Gaussians (Appendix \ref{appendix:inhibitory_functions}) unless otherwise stated. In the adding problem, always predicting the value 1 has mean square error $\approx .17$, and root mean square error (RMSE) $\approx .41$; a model has learned if it substantially improves on these baselines.
\begin{figure}
    \centering
    \begin{tabular}{c|c|c|c|c|c|c|c|c|c|c} \cline{2-10}
        \emph{input} & .2 & .6 & .9 & .1 & .4 & .4 & .7 & .1 & .8 & \emph{output} \\ \cline{2-10}
        \emph{mask}  & 0  &  0 &  1 &  0 &  1 &  0 &  0 &  0 &  0 & 1.3 \\ \cline{2-10}
    \end{tabular}
    \caption{The adding problem inputs (left) and intended output (right).}
    \label{fig:adding_problem}
\end{figure}

We tested a single layer RNN with a variety of moduli regularizers on this task to experiment on the manner in which moduli regularizers interact with long term memories. Moduli regularization broadly generalizes previous specialized architectures designed for the adding problem: for example, IndRNNs \cite{li2018independently} are equivalent to RNNs trained with moduli regularization, where the inhibitor function $f$ is given by $f(0) = 0,$ and $f(x) = \infty$ for all $x > 0$. 

\noindent {\bf Control experiments:} We repeat the control experiments used in the Navigation and control experiments. Table \ref{tab:adding_problem_control} shows that toroidal moduli regularizers outperform both $L_1$ regularizers and shuffled moduli regularizers at the 95\% sparsity threshold. 

\begin{table}[]
    \centering
    \caption{Root mean square error (standard deviation) over five experiments at 95\% sparsity. Toroidal moduli regularization results in the highest quality sparse model, though no lottery ticket sparse architectures are able to train.}
    \label{tab:adding_problem_control}
    \begin{tabular}{ccccc} \toprule
 & Torus & No reg & L1 & Permuted \\ \cmidrule{2-5}
 0\% Sparse RMSE & 0.040 (0.006) & 0.036 (0.006) & 0.036 (0.007) & 0.042 (0.010) \\
95\% Sparse RMSE & \textbf{0.111} (0.035) & 0.142 (0.122) & 0.265 (0.113) & 0.142 (0.038) \\
95\% Lottery RMSE & 0.404 (0.003) & 0.407 (0.001) & 0.409 (0.004) & 0.400 (0.010) \\ \bottomrule 
\end{tabular}
\end{table}

\noindent {\bf Trained embeddings:} Training the embedding allows the model to find superior regularizer arrangements for both the torus and $\BR^3$, as we show in Table \ref{tab:adding_problem_trained}. 

\begin{table}[]
    \centering
\begin{tabular}{lcccc} \toprule
 & Torus trained & Torus fixed & $\BR^3$ trained & $\BR^3$ fixed \\ \cmidrule(l{4pt}r{2pt}){4-5}
0\% Sparse RMSE & \textbf{0.035} (0.005) & 0.040 (0.006) & 0.044 (0.005) & \textbf{0.037} (0.005) \\
95\% Sparse RMSE & \textbf{0.055} (0.012) & 0.111 (0.035) &  \textbf{0.044} (0.012) & 0.050 (0.026) \\
95\% Lottery RMSE & 0.406 (0.004) & 0.404 (0.003) & 0.407 (0.002) & 0.406 (0.002) \\ \bottomrule 
\end{tabular}
    \caption{Root mean square error (standard deviation) over five experiments at 95\% sparsity. Training the embedding results in improvements in both toroidal and $\BR^3$ regimes.}
    \label{tab:adding_problem_trained}
\end{table}

\noindent {\bf Moduli spaces:} We compare the same moduli space hyperparameters as tested above in Table \ref{tab:adding_problem_moduli}. The torus, circle, and 6-torus perform well at 95\% sparsity, while the Klein bottle and sphere perform poorly. 

\begin{table}[]
    \centering
    \caption{Root mean square error (standard deviation) over five experiments at 95\% sparsity, with lottery ticket trained sparse architectures.}
    \label{tab:adding_problem_moduli}
    \begin{tabular}{cccccc} \toprule 
 & Torus & Circle & Sphere & Klein & 6-Torus \\ \cmidrule{2-6}
0\% Sparse RMSE & 0.040 (0.006) & 0.038 (0.005) & 0.042 (0.014) & 0.039 (0.007) & 0.037 (0.009) \\
95\% Sparse RMSE & 0.111 (0.035) & 0.097 (0.016) & 0.148 (0.114) & 0.214 (0.122) & \textbf{0.088} (0.006) \\
95\% Lottery RMSE & 0.404 (0.003) & 0.407 (0.002) & 0.404 (0.005) & 0.404 (0.004) & \textbf{0.353} (0.088) \\  \\ \bottomrule 
\end{tabular}

\end{table}

Our results show that moduli regularization is compatible with long term recall in RNNs, an important feature for implementation. Moreover, it is important as a sample use case where learned moduli spaces substantially outperform fixed ones.

\section{Discussion}\label{sec:discussion}
We introduce a regularization scheme for recurrent neural nets influenced by the underlying stable dynamics of the recurrent system, which can be initialized by the researcher by applying special domain-specific knowledge. We do this by embedding the recurrent neurons into a fixed metric space, and differentially penalizing weights according to their distance on this metric space. By embedding neurons into manifolds, we reparameterized their dynamics into sparse, geometrically organized structures. We then implemented our regularizer on navigation, natural language processing, and adding problem recurrent architectures, showing that it functions as a highly effective method of sparsification. 

Our method is distinguished from previous sparsification methods in that it creates relatively stable sparse models in the sense of the lottery ticket hypothesis \cite{frankle2018lottery, frankle2020linear}. Prior work in model sparsification has found that training sparse models is heavily dependent on the weight initialization \cite{frankle2018lottery, gale2019state}. 
Incorporating moduli regularization into sparsification alleviates these problems in some examples. In addition to creating superior sparse models, the underlying sparse architectures are \emph{stable}: new models, with randomized initializations, can be trained on previously learned sparse architectures to high fidelity. This opens the door to future work analyzing the limitations of different sparse architectures, and the possibility of training \emph{a priori} sparse models. Additionally, moduli regularization can be used as an exploratory technique to identify plausible candidates for manifolds underlying neural computations. 

A particularly interesting result of our experiments is that we demonstrate the importance of global weight organization in RNNs. The results of Table \ref{tab3:nav_moduli_comparison} are particularly incisive: the RNN is able to be effectively sparsified exactly when the moduli space is a good approximation of the underlying dynamics. 

In future work, we are particularly interested in studying multi-modular regularization systems, in which different neurons are embedded in different moduli spaces, as well as adapting our work to different neural architectures and use-cases. An important generalization of our work is a method which better combines manifold learning (such as \cite{fan2021manifold, gu2019learning, DBLP:journals/corr/HuangWG16, saez2024neural, skopek2020mixed}) with moduli regularization to dynamically choose optimal moduli spaces for a given network.

\begin{ack}
This work has been partially supported by the Army Research Laboratory Cooperative Agreement No W911NF2120186. 
\end{ack}
\newpage

\bibliography{mybibliography}

\newpage

\appendix
\section{Hardware, hyperparameters, architectures}\label{appendix:hyperparameters}
Navigation
numerical experiments were run on a NVIDIA RTX A4500 GPU with two Intel(R) Xeon(R) Gold 6246R CPU @ 3.40GHz (32 processing cores) with 640GB of memory and 28GB of swap space. The natural language processing numerical experiments were run on a NVIDIA Tesla K80 GPU with two Intel Xeon E5-2637 CPUs @ 3.50GHz (8 processing cores) with 64GB of memory and 28GB of swap space. Each navigation experiment took $\approx 2.26$ hours, and each NLP experiment $\approx 2.69$ hours. Our code is available at \url{https://github.com/mackeynations/Moduli-regularizers}

\subsection{Navigation RNN}\label{appendix:rnn_hyperparameters}
We use the same hyperparameters and architecture as \cite{sorscher2023unified}, as listed below in Table \ref{tab3:nav_hypers}, with the exception of increasing the batch size and decreasing the number of training epochs. We also remove the $L_2$ regularization that was applied in \cite{sorscher2023unified}, since it substantially slows training. The code is freely available. 
\begin{table}[h!]
\caption{Hyperparmaeters for navigation experiments (Section \ref{sec3:navigation}).}
\begin{center}
\begin{tabular}{cc} \toprule 
 \textbf{Hyperparameters}    & \textbf{Values} \\ \hline 
  Optimizer   & Adam \cite{kingma2014adam} \\ 
  Batch size & 200 routes \\
  Batches tested & 30 000 \\
  Sequence length & 50 \\
  Learning rate & $10^{-4}$ \\
  Input \& output dim & 512 \\
  Hidden dim & 4096 \\ 
  Bias & No \\
  Nonlinearity & ReLU \\ 
  Regularizing factor & $10^{-5}$ \\
  Loss & Cross entropy \cite{mannor2005cross} of softmax of Gaussians \\
  Region size & 2.2 m square box \\ \bottomrule
\end{tabular}
\end{center}
\label{tab3:nav_hypers}
\end{table}

\textcolor{blue}{$L_1$ regularizers used the same regularizing factor as moduli regularizers. }

As all data was freely generated, there was no train/test split of the data: every tested path was unique.

The RNN is composed of:
\begin{enumerate}
    \item A linear encoder for the original landmark positions
    \item A linear encoder for velocity vectors
    \item The hidden update matrix
    \item A linear decoder from the hidden state to the relative landmark positions. 
\end{enumerate}
To calculate the loss, a softmax function
\begin{equation}\label{eq:softmax}
    \on{softmax}([x_0, x_1, ..., x_n]) = \left[\frac{e^{x_0}}{\sum_i e^{x_i}}, \frac{e^{x_1}}{\sum_i e^{x_i}}, ..., \frac{e^{x_n}}{\sum_i e^{x_i}}\right]
\end{equation}
is applied to the decoded values, which we call the predicted scores. The underlying true scores are given by $[y_0, ..., y_n] = \on{softmax}([g_0, ..., g_n])$, where $g_i = \frac{1}{12\sqrt{2\pi}}e^{-d_i^2/(2*12^2)}$, where $d_i$ is the distance from the final position to landmark $i$. We then compare the predicted scores to the true scores using cross entropy loss:
\begin{equation}\label{eq:cross_entropy}
    \on{crossentropy}([x_i]_{i \le n}, [y_i]_{i \le n}) = - \sum_{i=0}^n x_i \log y_i.
\end{equation}

\begin{figure}
    \centering
    \includegraphics[width=0.95\linewidth]{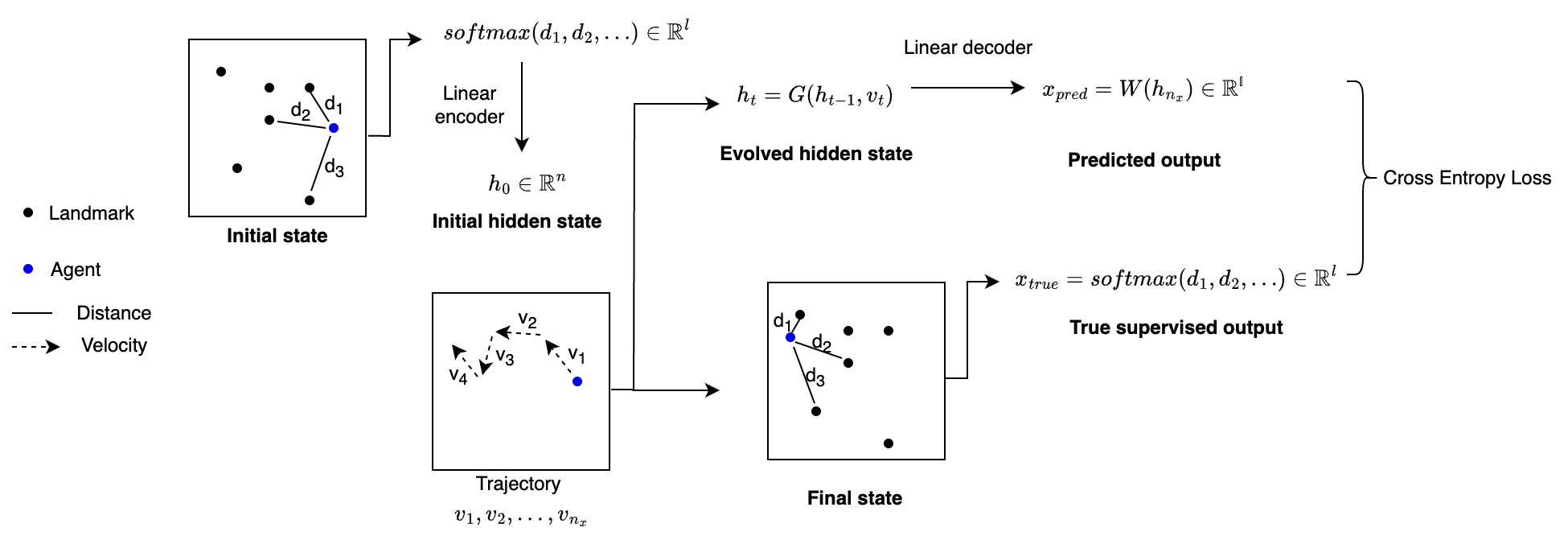}
    \caption{Illustration of the navigation problem and architecture. }
    \label{fig:appendix_navigation_explainer}
\end{figure}

The input and output functions of the RNN and the focus on landmarks is not as strange as it might appear. A very similar, though much larger, architecture was used in \cite{mirowski2018streetnav} for navigation in large cities based purely on visual input. The rationale is that introduction of new landmarks is arbitrarily extensible, lending it greater potential generalizability of the system. We depict a diagram of the full architecture in Figure \ref{fig:appendix_navigation_explainer}, and in shorter form in Figure \ref{fig:navigation_explainer}.

We ran a number of experiments whose results are not reported in the paper, including adding regularizers to other model parameters, and exploratory work to discover appropriate percentiles for sparsification, for which compute times were not recorded. 

%

\subsection{NLP RNN}\label{appendix:word_architecture}
We use the recommended hyperparameters of and architecture of \cite{pytorch_examples}, as listed below in Table \ref{tab3:nlp_hyps}. The code is available under the BSD 3-Clause License. For training and testing data, we use the Wikitext-2 dataset of high quality Wikipedia articles \cite{merity2016pointer}, available under Creative Commons Attribution Share Alike 3.0 license.
\begin{table}[h!]
\caption{Hyperparmaeters for NLP experiments (Section \ref{sec3:nlp}).}
\begin{center}
    \begin{tabular}{cc} \toprule 
        {\bf Hyperparameters} & {\bf Values} \\ \hline 
        Optimizer & SGD \\ 
        Nonlinearity & $\tanh$ \\
        Number of recurrent layers & 3 \\ 
        Bias & Yes \\
        Batch size & 20 \\ 
        Embedding dim & 650 \\
        Hidden dim & 650 \\ 
        Dropout & 0 \\ 
        Starting learning rate & 20 \\
        Learning rate decay & .25 \\ 
        Gradient clipping threshold & .25 \\ 
        Backprop through time length & 35 \\ \bottomrule 
    \end{tabular}
\end{center}
\label{tab3:nlp_hyps}
\end{table}

\textcolor{blue}{$L_1$ regularizers used the same regularizing factor as moduli regularizers. }

The data was split as in the Wikitext-2 dataset \cite{merity2016pointer}, which corresponds to roughly 82\% of data reserved for training, training, 8\% for validation, and 10\% for test. All reported results are on the test dataset.

The RNN is composed of:
\begin{enumerate}
    \item An encoder from the list of words to $\BR^{650}$
    \item Three recurrent neural layers.
    \item A decoder from neural output to a one-hot encoding of the list of words.
\end{enumerate}
As a loss function, we use the cross entropy (Equation \eqref{eq:cross_entropy}) of the softmax (Equation \eqref{eq:softmax}) of the decoder output, as compared with the distribution which assigns probability 1 to the true next word, and 0 to all other words.

We additionally ran exploratory experiments to identify good thresholds for sparsification, for which we did not record compute time. The interested reader is recommended Tu's (2011) book \cite{tu2011manifolds}.

\begin{thisnote}

\subsection{Adding problem RNN}\label{appendix:adding_problem_hypers}
We used the hyperparameters shown in Table \ref{tab:adding_problem_hypers}.
\begin{table}[h!]
    \centering
    \caption{Hyperparameters for the adding problem experiments (Section \ref{sec:adding_problem_experiments}).}\label{tab:adding_problem_hypers}
    \begin{tabular}{cc}\toprule 
       \textbf{Hyperparameters}  & \textbf{Values}  \\ \hline 
       Optimizer  & Adam \cite{kingma2014adam} \\ 
       Batch size & 32 \\
       Batches tested & 500,000 \\ 
       Sequence length & 50, or as specified \\ 
       Learning rate & $10^{-4}$ \\ 
       Hidden dim & 128 \\ 
       Bias & Yes \\ 
       Nonlinearity & ReLU \\ 
       Regularizing factor & $10^{-3}$ \\ 
       Loss & Mean square error \\ 
       Grad clip & 0.5
    \end{tabular}
    
\end{table}

$L_1$ regularizers used the same regularizing factor as moduli regularizers.

\end{thisnote}

\subsection{Speech recognition}\label{appendix:speech_hypers}
\begin{thisnote}

We use the same hyperparameters as \cite{deepspeech2}, as shown in Table \ref{tab:speech_hypers}, with the exception of our use of a single layer RNN, rather than a multi-layer bidirectional GRU. 
\begin{table}[h!]
    \centering
    \caption{Hyperparameters for speech recognition experiments.}\label{tab:speech_hypers}
    \begin{tabular}{c|c}
        {\bf Hyperparameters} & {\bf Values}  \\ \hline
        Optimizer & AdamW \cite{loshchilov2017decoupled} \\ 
        Nonlinearity & tanh (recurrent), GELU \cite{hendrycks2016gaussian} (non-recurrent) \\
        Number of hidden layers & 1 \\ 
        Bias & Yes \\
        Number of convolutional layers & 3 \\ 
        Convolutional channels & 32 \\
        Convolutional kernel & $3 \times 3$, with stride 1 \\
        Batch size & 20 \\ 
        Hidden dim & 512 \\ 
        Dropout & 0.1 \\ 
        Learning rate & $5 \times 10^{-4}$ \\ 
        Learning rate annealing strategy & Linear \\
        Epochs & 10 \\
        Regularizing constant & $10^{-2}$ \\ 
        Loss & CTCLoss \cite{graves2006connectionist}
    \end{tabular}
\end{table}

$L_1$ regularizers used the same regularizing factor as moduli regularizers.

The data was split as follows: 80\% training, 20\% testing. The Librispeech dataset \cite{panayotov2015librispeech} is available under the Creative Commons 4.0 license. 

Based on the Deep Speech 2 model \cite{deepspeech2}, the RNN structure is given by:
\begin{enumerate}
    \item Compute the Mel spectogram \cite{shen2018natural} of the data
    \item Three residual convolutional layers process the spectogram
    \item Three bidirectional RNNs process the output of the convolutional layers 
    \item Two linear layers with GELU nonlinearity
\end{enumerate}
We then used the Connectionist Temporal Classification Loss (CTCLoss) \cite{graves2006connectionist}, a paradigm designed for variable many-to-one mappings, as a loss function.

\end{thisnote}

\section{Manifolds}\label{appendix:manifolds}
A manifold is a topological space which locally is isomorphic to the Euclidean space \cite{hatcher2005algebraic}. A sphere like the earth's surface, for example, has global curvature: it isn't isomorphic to $\BR^2$. However, a person walking on the earth could easily mistake it as flat: a square grid is a fine way to organize something as large as a city. This motivates the following definitions:

\begin{defn}
    A continuous map $f: X \to Y$ is a homeomorphism if it is bijective, and its inverse $f^{-1}: Y \to X$ is continuous. We say $X$ and $Y$ are homeomorphic if there exists a homeomorphism $f: X \to Y$. 
\end{defn}
\begin{defn}
    An $n$-dimensional manifold is a topological space for which there exists an open cover $\{ U_i \}_{i \in I}$ such that $U_i$ is homeomorphic to $\BR^n$ for all $i \in I$. 
\end{defn}

Manifolds are particularly important because it is not possible to perform calculus on general topological spaces. Since manifolds are locally parameterized by copies of $\BR^n$ though, it is sometimes possible to perform calculus on them. This is possible whenever the homeomorphisms $U_i \cong \BR^n$ are compatible with one another---these are known as $\mathcal{C}^k$ manifolds, $0 \le k \le \infty$. $\mathcal{C}^\infty$ manifolds are also known as smooth manifolds. \begin{thisnote}
    A manifold is $\mathcal{C}^k$ if, for each pair of elements of the open cover $U_i, U_j$, with $f_i: U_i \to \BR^n$, $f_j: U_j \to \BR^n$ homemorphisms, the functions $f_i \circ f_j^{-1}$ and $f_j \circ f_i^{-1}: \BR^n \to \BR^n$ are $k$-times differentiable. 
\end{thisnote}

We particularly care about compact manifolds. For manifolds that are embedded in Euclidean space (for example, a sphere or torus in three dimensional space), these are spaces which contain all their limit points and are bounded. Compact spaces have the nice property that there is an upper bound on how far away pairs of points can be. \textcolor{blue}{Common examples of manifolds include the circle, the sphere, and the torus. A figure 8, however, is not a manifold: at the intersection point of the two loops it is not homeomorphic to the line. }


\section{Covering maps}\label{appendix:covering_spaces}

Covering maps are a special class of highly controlled functions. 
\begin{defn}
    A continuous map $f: X \to Y$ is a covering map if it is surjective and for every point $y \in Y$, there is an open set $U \subseteq Y$ containing $y$ such that
    \[ f^{-1}(U) \cong \coprod_{d \in D} U,\]
    where $D$ is a discrete set. We say $X$ is a covering space for $Y$. 
\end{defn}

For example, the map $p: \BR^1 \to S^1$ from the real line to the unit circle given by
\[ x \mapsto (\cos(x), \sin(x)) \]
is a covering space. \textcolor{blue}{We illustrate this example in Figure \ref{fig:covering_space_example}.}

\begin{figure}
\centering
\begin{tikzpicture}
\begin{axis} [
    view={0}{30},
    axis lines=none,
    ymin=-2,
    ymax=5,
    xmin=-2,
    xmax=2]

    \addplot3 [thick, <->, blue, domain=3:7*pi, samples = 100, samples y=0] ({sin(deg(-x))}, {cos(deg(-x))}, {x});
    \addplot3 [thick, black , domain=0:2*pi, samples = 100, samples y=0] ({sin(deg(x))}, {cos(deg(x))}, -3);
    \addplot3 [thick, only marks, blue, mark=...] ({sin(deg(-3))}, {cos(deg(-3)}, {3});
\end{axis}
\end{tikzpicture}
\caption{\textcolor{blue}{$\BR^1$, represented as a blue line, is a covering space for $S^1$, represented as the black circle. The projection map is $x \mapsto (\cos x, \sin x)$. }} \label{fig:covering_space_example}
\end{figure}

\begin{lem}
    If $\wt{X} \xrightarrow{f} X$, $\wt{Y} \xrightarrow{g} Y$ are covering maps, then $\wt{X} \times \wt{Y} \xrightarrow{f \times g} X \times Y$ is a covering map.
\end{lem}
\begin{proof}
    For any point $(x, y) \in X \times Y$, let $U_x \ni x$, $U_y \ni y$ be open sets in $X$ and $Y$ such that $f^{-1}(U_x) \cong \coprod_{d \in D} U_x, g^{-1}(U_y) \cong \coprod_{d' \in D'} U_y$. Then $(f \times g)^{-1}(U_x \times U_y) \cong \coprod_{(d, d') \in D \times D'} U_x \times U_y$. 
\end{proof}

As a result, there is a covering map $p \times p: \BR^2 \to S^1 \times S^1$ from two dimensional Euclidean space to the torus.  There is a special kind of covering map, called a universal covering map. This is a covering space $X \to Y$ where $X$ has trivial fundamental group: that is, all loops in the space can be filled in. The interested reader is recommended Hatcher's (2005) book \cite{hatcher2005algebraic} for more details.

\section{Regularizer Parameterizations}\label{appendix:smooth_structures}
To extract metric properties from the manifolds used for moduli regularizers, it is necessary to fix a parameterization. We used the following parameterizations in our experiments:
\begin{itemize}
    \item The circle: the locus of $(x, y)$ such that $x^2 + y^2 = 25$.
    \item The sphere: The locus of $(x, y, z)$ such that $x^2 + y^2 + z^2 = 25$.
    \item The torus: $\BR^2/10\BZ^2$.
    \item The Klein bottle: $[0, 10] \times [0, 10]/\sim$, where $\sim$ is the equivalence relation which identifies the points $(x, 0) \sim (x, 10)$ and $(0, y) \sim (10, 10-y)$.
    \item The 6-torus: $\BR^6/10\BZ^6$. 
\end{itemize}
To create embeddings $\{1, 2, ..., n\} \to \mathcal{M}$ of hidden state neurons into each manifold, we sampled points independently from the uniform distribution on $\mathcal{M}$ for each $\mathcal{M}$.

\section{Inhibitor functions}\label{appendix:inhibitory_functions}
\begin{figure}
    \centering
    \includegraphics[width=.9\linewidth]{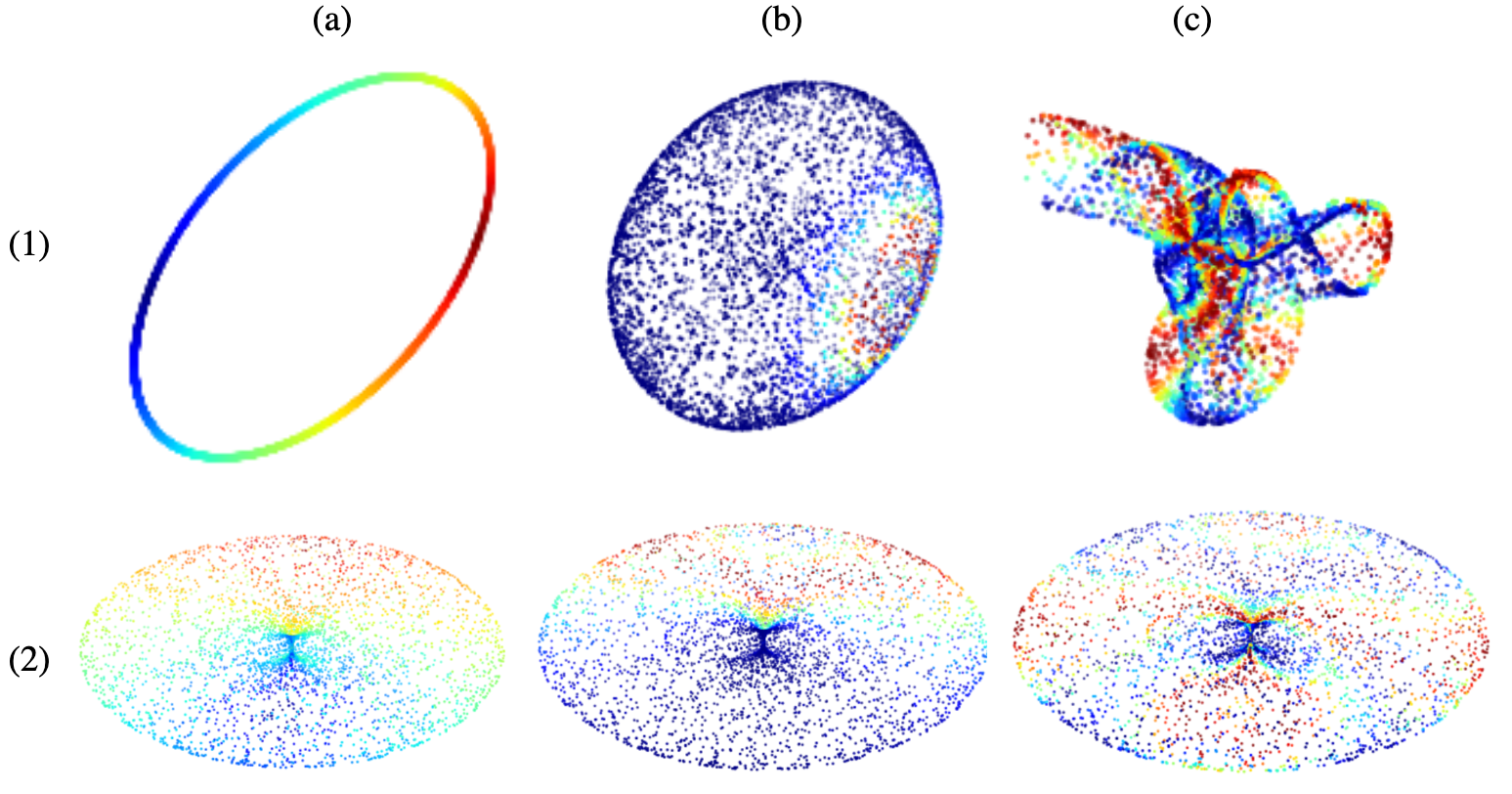}
    \caption{\textcolor{blue}{We illustrate the manner in which different moduli spaces and different inhibitor functions induce different weights in the hidden update matrix. A neuron $y$ is fixed. Red points depict output neurons $z$ with low regularizing values, and therefore potentially large weights $w_{yz}$, while blue points are output neurons $z'$ with high regularizing values, and therefore smaller potential weights $w_{yz'}$. Row (1) depicts the circle, sphere, and Klein bottle, while row (2) depicts the torus. Column (a) has inhibitor function $f(d) = d^2$, column (b) has inhibitor function $f(d) = 10 - 10(e^{-\frac{d^2}{5}} -e^{-d^2})$, and column (c) has inhibitor function $f(d) = 10 + 10\cos(2d)$. }}
    \label{fig3:inhibitors}
\end{figure}

Here we provide additional data on different inhibitor functions in our experiments. For this comparison, we use the torus as a moduli space. Here $c, \sigma, \sigma_1, \sigma_2$, and $\mu$ are positive constants. Difference of Gaussians (DoG) and Sinuosoid inhibitor functions can be natively used with trained embeddings, while Diffusion cannot without additional regularizing terms. We test using the following inhibitory functions:
\begin{itemize}
    \item Difference of Gaussians (DoG): $f(x) = c - c(e^{-\frac{x^2}{\sigma_2}} -e^{-\frac{x^2}{\sigma_1}})$
    \item Diffusion: $f(x) = cx^n$
    \item Sinuosoid: $f(x) = c + c\cos{\mu x}$
\end{itemize}
Results are shown in Tables \ref{tab3:nav_inhibitors} and \ref{tab3:word_inhibitors} for navigation and natural language processing, respectively. We illustrate how these inhibitor functions appear in Figure \ref{fig3:inhibitors}.

\begin{table}[htb!]
    \centering
    \caption{Error in cm (standard deviation) of navigation RNNs (Section \ref{sec3:navigation}). The difference of Gaussians inhibitory function induces mildly superior results, as compared to alternatives}
\begin{tabular}{ccc} \toprule
 & DoG & Diffusion \\ \cmidrule{2-3}
 75\% Sparse Err. & 4.86 (0.30) & 4.66 (0.09) \\
75\% Lottery Err. & 7.28 (0.69) & 7.05 (0.57) \\
90\% Sparse Err. & 5.69 (0.43) & 14.72 (17.43) \\
90\% Lottery Err. & 12.49 (2.03) & 11.32 (0.62) \\
\bottomrule
\end{tabular}
    \label{tab3:nav_inhibitors}
\end{table}

\begin{table}[htb!]
    \centering
\caption{\textcolor{blue}{Perplexity} \textcolor{red}{Accuracy} (standard deviation) of next-word predictor RNNs (Section \ref{sec3:nlp}). The difference of Gaussians inhibitor provides substantially better results, both during sparsification and in lottery ticket experiments.}
\begin{tabular}{cccc} \toprule
 & DoG & Diffusion & Sinuosoid \\ \cmidrule{2-4}
\textcolor{blue}{0\% Sparse} & 69.26 (16.31) & 60.49 (11.23) & 130.96 (27.59) \\
90\% Sparse & 43.41 (7.53) & 74.65 (38.75) & 112.46 (38.98) \\
90\% Lottery & 54.80 (8.70) & 81.91 (27.02) & 113.77 (0.50) \\
98\% Sparse & 55.80 (13.96) & 48.75 (10.42) & 103.11 (31.08) \\
98\% Lottery & 61.44 (6.06) & 75.71 (9.19) & 114.57 (1.24) \\ \bottomrule
\end{tabular}
    \label{tab3:word_inhibitors}
\end{table}


\section{Heatmap for trained navigation RNN}\label{appendix:heatmap}
\begin{figure}[h]
    \centering
    \includegraphics[width=.45\linewidth]{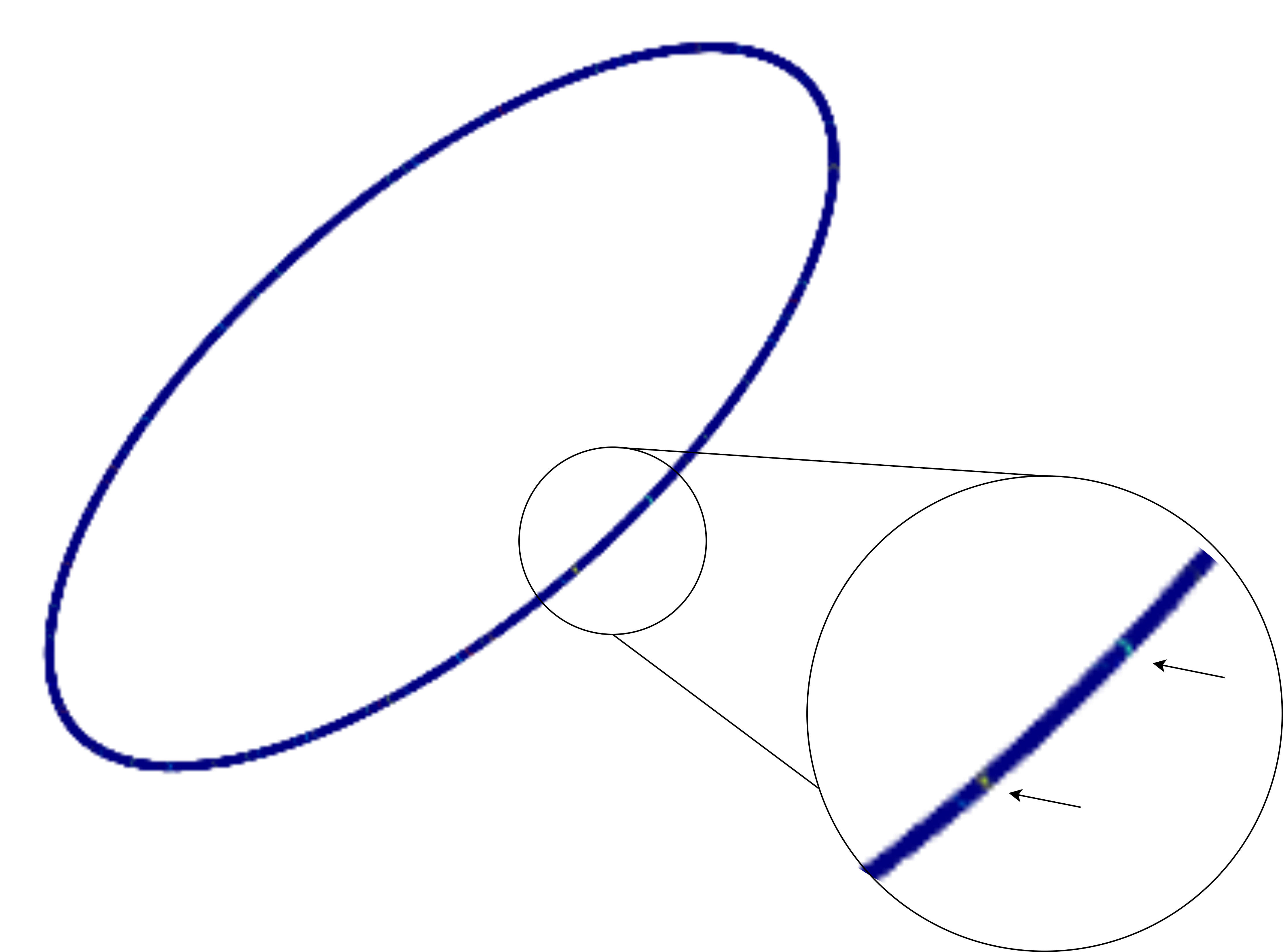}
    \caption{Heatmap of neural weights in a navigation RNN.}
    \label{fig3:heatmap_circle}
\end{figure}

We show a sample heatmap of the connections of one hidden neuron in our navigation RNN (Section \ref{sec3:navigation}) with moduli space the circle and inhibitor function the difference of Gaussians (Equation \eqref{eq:DoG}) in Figure \ref{fig3:heatmap_circle}. 

\section{Sensitivity to the weight decay factor}\label{appendix:ablation}
The influence of regularization on trained models is controlled by the weight decay constant it is multiplied by ($\lambda$ in Equation \eqref{eq:regularized_loss_func_general}). We show the results of varying the regularizing coefficient $\lambda$ for our navigation experiments in Tables \ref{tab:ablation_results} and \ref{tab:ablation_results2} and the regularizing coefficient $\lambda$ for NLP in Tables \ref{tab3:nlp_ablation} and \ref{tab3:nlp_ablation2}. Due to the computational costs, we study both cases only at $90\%$ sparsity, and at 10 different weight decay values. Due to a typo during execution, navigation experiments and NLP experiments occurred at slightly different weight decay values---we decided that the results were sufficiently uniform that it was not worth rerunning the experiments for precisely equal comparisons. All results here record different trials than the those presented in the main paper, even when hyperparameters are identical in the two settings. 

\begin{table}[h]
\centering 
\begin{tabular}{lccccc} \toprule
 & $1$ & $10^{-1}$ & $10^{-2}$ & $10^{-3}$ & $10^{-4}$ \\ \cmidrule{2-6}
$L_1$ 90\% Sparse&  62.37 (12.91) & 58.93 (13.95) & 60.49 (17.51) & 73.93 (9.47) & 70.15 (3.17) \\
Torus 90\% Sparse & 41.75 (0.49) & 9.13 (1.12) & 17.88 (6.24) & 37.52 (15.22) & 58.16 (15.21) \\
Shuffled 90\% Sparse & 72.13 (1.72) & 23.12 (1.67) & 16.92 (5.88) & 31.87 (10.94) & 70.09 (2.02) \\
\bottomrule
\end{tabular}
\caption{We show sensitivity of the sparsification of our navigation experiments at 90\% sparsity of $L_1$ regularization, moduli sparsification on the torus, and shuffled moduli regularizers. Here we report mean error (cm) (standard deviation) over five trials.}
\label{tab:ablation_results}
\end{table}

\begin{table}[h]
\centering 
\begin{tabular}{lccccc} \toprule
 & $6$ & $6 \times 10^{-1}$ & $6 \times 10^{-2}$ & $6 \times 10^{-3}$ & $6 \times 10^{-4}$ \\ \cmidrule{2-6}
$L_1$ 90\% Sparse & 70.74 (6.52) & 60.01 (17.62) & 64.73 (18.85) & 74.91 (6.08) & 57.54 (17.36) \\
Torus 90\% Sparse & 47.30 (0.49) & 38.16 (1.39) & 9.22 (1.92) & 12.47 (0.97) & 62.66 (17.92) \\
Shuffled 90\% Sparse & 58.96 (5.36) & 59.58 (7.99) & 18.03 (1.99) & 15.45 (3.64) & 58.69 (14.66) \\
\bottomrule
\end{tabular}
\caption{We show extended sensitivity data of the sparsification of our navigation experiments at 90\% sparsity of $L_1$ regularization, moduli sparsification on the torus, and shuffled moduli regularizers. Here we report mean error (cm) (standard deviation) over five trials.}
\label{tab:ablation_results2}
\end{table}

\begin{table}[h]
    \centering
    
\begin{tabular}{lccccc} \toprule
 & 1 & $10^{-1}$ & $10^{-2}$ & $10^{-3}$ & $10^{-4}$ \\ \cmidrule{2-6}
$L_1$ 90\% Sparse & 106.04 (28.66) & 118.17 (3.65) & $179.53^*$ (99.71) & 117.98 (2.73) & 101.50 (15.24) \\
Torus 90\% Sparse & 114.64 (26.32) & 50.46 (12.08) & 63.04 (35.18) & 90.79 (35.60) & 124.77 (16.44) \\
Shuffled 90\% Sparse & 85.93 (38.53) & 57.14 (18.43) & 81.63 (40.31) & 133.23 (26.05) & 103.54 (27.70) \\ \bottomrule
\end{tabular}
\caption{We show sensitivity of the sparsification of our NLP experiment at 90\% sparsity of $L_1$ regularization, moduli sparsification on the torus, and shuffled moduli regularizers. Here we report perplexity of the next-word prediction of a single trial. $\,^*$ marks the exclusion of one trial which suffered representation collapse.}
    \label{tab3:nlp_ablation}
\end{table}

\begin{table}[h]
\centering
\begin{tabular}{lccccc} \toprule
 & 5 & $5 \times 10^{-1}$ & $5 \times 10^{-2}$ & $ 5 \times 10^{-3}$ & $5 \times 10^{-4}$ \\ \cmidrule{2-6}
$L_1$ 90\% Sparse & 118.47 (2.62) & 106.03 (24.47) & 120.29 (2.63) & 108.85 (31.30) & 184.67 (132.10) \\
Torus 90\% Sparse & 111.88 (23.33) & 66.47 (21.26) & 43.63 (7.80) & 103.55 (28.41) & 117.96 (1.83) \\
Shuffled 90\% Sparse & 110.85 (23.56) & 62.17 (13.80) & 62.47 (7.05) & 162.42 (124.24) & 117.39 (16.86) \\ \bottomrule
\end{tabular}
\caption{We show extended sensitivity data of the sparsification of our NLP experiment at 90\% sparsity of $L_1$ regularization, moduli sparsification on the torus, and shuffled moduli regularizers. Here we report perplexity of the next-word prediction of a single trial. }
    \label{tab3:nlp_ablation2}
\end{table}

\section{Trained embeddings to learn manifolds}\label{appendix:trained_persistence}

\begin{figure}[h]
    \centering
    \includegraphics[width=.7\linewidth]{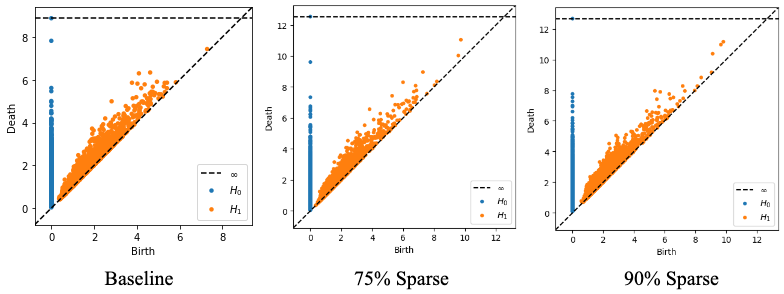}
    \caption{Persistence homology applied to neurons with trained embeddings in $\BR^3$ are indistinguishable from random points.}
    \label{fig:tda}
\end{figure}

To test for manifold learning with trained embeddings, we apply persistence homology \cite{edelsbrunner2008persistent} to navigation experiments (Section \ref{sec3:navigation}) with moduli space $\BR^3$ and difference of Gaussians inhibitor function (Equation \eqref{eq:DoG}) shown in Table \ref{tab:trained_r3}. The persistence diagrams for three of the neural point cloud are shown in Figure \ref{fig:tda}. There is no clear learning of a manifold present: the neurons appear randomly distributed. For these experiments, we initialized neurons on a uniform distribution in $[0, 10]^3 \subset \BR^3$.

\begin{thisnote}
\section{Moduli regularization for non-recurrent architectures}\label{appendix:non_recurrent_layers}
\subsection{Fully connected layers}
Given a fully connected layer of a neural net, no changes to Eq. \ref{eq:moduli_reg} are necessary to create a moduli-like regularizer: the hidden weight update matrix is a special case of a fully connected layer. However, as neurons in layer $k$ of an MLP are not associated with neurons in layer $k+1$, different embeddings from the set of neurons into the manifold $\iota_k: [n_k] \hookrightarrow \mathcal{M} \hookleftarrow [n_{k+1}] :\iota_{k+1}$ should be chosen. 

\subsection{Convolutional layers}
Regularization and sparsification in convolutional layers of neural nets has historically been unsuccessful \cite{NEURIPS2022_sparsity}. The non-local nature of convolutions causes them to be highly resilient to overfitting, obviating particularly the need for regularization. We describe a generalization of moduli regularization to general convolutional layers in Section \ref{subappendix:moduli_for_conv}. Our experiments broadly comported with past results, however: adding moduli regularization to convolutional layers of neural nets results in mildly worse performance, in terms of both training time and validation loss. Nevertheless, this generalization is vital for implementation of moduli regularization in convolutional recurrent neural nets (CRNNS), which we will explore in future work. 

The paper \cite{topdeeplearning} implements something very similar: the primary differences to our paper are that they hard-code both weights and sparse structures, rather than regularize their neural nets, and they only consider the Klein bottle and the circle in the second convolutional layer of a neural net. They found some positive results when testing with class correlated noise

\subsubsection{Moduli regularization for convolutional layers}\label{subappendix:moduli_for_conv}
For linear/fully connected layers, moduli regularization was implemented by embedding neurons into a manifold $\CM$ and penalizing weights between neurons that are very distant. The analogue for convolutional layers is to embed channels into a manifold and penalize weights between channels that are very distant. In equations, the regularization term is
\[ \sum_{c, c' \in \text{Channels}} d_{\mathcal{M}}(i(c), i(c'))|T_{c,c'}|^2, \]
where $i$ is a chosen embedding of the set of channels into $\CM$, $d_{\CM}$ is a distance function on $\CM$, $T_{c,c'}$ is convolved grid between $c$ and $c'$ (or 0 if $c'$ is not in a layer immediately subsequent to that of $c$), and $|-|^2$ is the $L^2$ norm on the vector space. 

\end{thisnote}



\end{document}